\documentclass{article}
\usepackage{arxiv}
\usepackage[T1]{fontenc}
\usepackage{graphicx}
\usepackage{enumitem}
\usepackage{amsfonts}
\usepackage{subcaption}
\usepackage{comment}
\usepackage{booktabs}
\usepackage{pdfpages}
\usepackage{tabularx, multirow}
\usepackage{orcidlink} 
\usepackage{wrapfig}
\usepackage{todonotes}
\usepackage[many]{tcolorbox}

\definecolor{main}{HTML}{b2beb5}
\definecolor{sub}{HTML}{ffffff} 
\tcbset{
    sharp corners,
    colback = white,
    before skip = 0.2cm,    
    after skip = 0.5cm      
}  
\newtcolorbox{boxH}{
    colback = sub, 
    colframe = main, 
    boxrule = 0pt, 
    leftrule = 6pt 
}
\newtcolorbox{boxD}{
    colback = sub, 
    colframe = main, 
    boxrule = 0pt, 
    toprule = 3pt, 
    bottomrule = 3pt 
}
\usepackage{amsmath}
\newcommand{\etal}{\textit{et al}. } 
\DeclareMathOperator*{\argmax}{arg\!\max}
\DeclareMathOperator*{\argmin}{arg\!\min}

\usepackage[belowskip=2pt,aboveskip=2pt]{caption}

\makeatletter
\newcommand\footnoteref[1]{\protected@xdef\@thefnmark{\ref{#1}}\@footnotemark}
\makeatother

\usepackage{mwe}
\usepackage{wrapfig}

\begin{document}

\renewcommand{\headeright}{}
\renewcommand{\undertitle}{(Accepted at FPS 2025)}
\renewcommand{\shorttitle}{Watch Out for the Lifespan: Evaluating Backdoor Attacks Against Federated Model Adaptation}

\title{Watch Out for the Lifespan: Evaluating Backdoor Attacks Against Federated Model Adaptation}

\date{}

\author{ {Bastien Vuillod, Pierre-Alain Moëllic} \\
	CEA-Leti, Centre CMP, Equipe Commune CEA-Leti - Mines Saint-Etienne, F-13541 Gardanne, France \\
	Univ. Grenoble Alpes, CEA-Leti, F-38000 Grenoble, France\\
	\texttt{\{name\}.\{surname\}@cea.fr} \\
	\And
	{Jean-Max Dutertre} \\
	Mines Saint-Etienne, CEA, Leti, Centre CMP,\\ F-13541 Gardanne France\\
	\texttt{dutertre@emse.fr} \\
}

\maketitle              

\begin{abstract}
Large models adaptation through Federated Learning (FL) addresses a wide range of use cases and is enabled by Parameter-Efficient Fine-Tuning techniques such as Low-Rank Adaptation (LoRA). However, this distributed learning paradigm faces several security threats, particularly to its integrity, such as backdoor attacks that aim to inject malicious behavior during the local training steps of certain clients. We present the first analysis of the influence of LoRA on state-of-the-art backdoor attacks targeting model adaptation in FL. Specifically, we focus on backdoor lifespan, a critical characteristic in FL, that can vary depending on the attack scenario and the attacker's ability to effectively inject the backdoor. A key finding in our experiments is that for an optimally injected backdoor, the backdoor persistence after the attack is longer when the LoRA's rank is lower. Importantly, our work highlights evaluation issues of backdoor attacks against FL and contributes to the development of more robust and fair evaluations of backdoor attacks, enhancing the reliability of risk assessments for critical FL systems. Our code is publicly available.

\end{abstract}

\setcounter{footnote}{0} 
\section{Introduction}
\label{introduction}
Federated Learning (FL) is a distributed learning paradigm that allows clients to collaborate in training a model using only their own private data \cite{mcmahan2017communication}.  
More recently, FL has emerged as a relevant solution for adapting pre-trained large models to downstream tasks, which is now a major topic in Machine Learning. Although these models are large-scale ones (e.g., ViT, LLM), 
local training at the client level is made possible by now-standard techniques of Parameter-Efficient Fine-Tuning (PEFT), which significantly reduce the number of trainable parameters. Among them, Low-Rank Adaptation, (LoRA) \cite{hu2021lora} has demonstrated remarkable performance, and many variations have been proposed in both centralized \cite{mao2025survey,meng2025pissa} and FL contexts \cite{sun2024improving,wangflora}.

While FL attracts considerable interest as a distributed paradigm, it is not without major security challenges including 
threats targeting the availability and integrity of FL systems. One or more malicious users can severely disrupt the average performance of the global model (Byzantine attacks \cite{baruch2019little}), or inject a targeted malicious behavior by modifying the local model and/or poisoning their training data. This latter scenario corresponds to one of the most extensively studied threats: backdoor attacks, which are already a major concern in the centralized learning paradigm. However, attacking a FL system with backdoor attacks presents distinct challenges. In typical scenarios, the attacker manipulates a limited number of clients, usually within a constrained time frame, and the poisoned models are eventually diluted during the aggregation process on the central server. As a result, a robust and fair evaluation of backdoor attacks in FL is complex and raises numerous methodological questions which are at the core of this work. More specifically, beyond the static performance of the \textit{backdoor injection}, its \textit{durability} throughout the federated process has gradually become an important notion in the state-of-the-art \cite{zhang2023a3fl,zhang2022neurotoxin}. Moreover, most studies on backdoor attacks in FL focus on standard training-from-scratch scenarios with classical CNN models. To our knowledge, the impact of backdoor attacks in FL for large model adaptation 
using PEFT techniques remains unexplored, despite the growing importance of such use cases in modern ML applications.

\paragraph{\textbf{Contributions.}} Our contribution can be listed as follow:
\begin{itemize}
    \item We first discuss the complexity of evaluating attacks in FL and highlight the importance of some system parameters, attack scenarios, and differences between backdoors attacks in FL.
    \item Then, our main contribution is an in-depth analysis of the impact of LoRA on backdoor attacks in the context of large model adaptation via FL. More specifically, the impact both on backdoor injection and lifespan.
    \item In light of these dynamics, we discuss a method for reducing the backdoors lifespan
and provide requirements for improving the evaluation of backdoors in FL, along with directions for future work. 
    \item Despite clear hypothesis of the threat scenario, the study of the learning dynamics with different attack settings sheds light on the complexity of evaluating backdoor attacks in FL and questions the ranking of the attacks when evaluated in new contexts.
\end{itemize}   
 Our code and additional results are available at \url{https://gitlab.emse.fr/securityml/lora_backdoor_fl}.

\section{Background and Related Work}
\label{related_work}
\paragraph{\textbf{Federated Learning.} } We classically formalize a FL system as a set $\mathcal{C}$ of $N$ clients connected to a server $S$ responsible for the aggregation. Note that, in the case of cross-device FL, the number of clients can vary significantly, ranging from a few units (e.g., industrial IoT) to several thousand devices (e.g., smartphones). Each client $k \in 0,1,...,N-1$ trains a local model on their own dataset $\mathcal{D}_k$. The overall training dataset is noted as $\mathcal{D}=\cup_{k=0}^{N-1}\mathcal{D}_k$. At every communication round $t$, a subset of clients $\mathcal{P}_t \subset \mathcal{C}$ is randomly selected by $S$ which sends them the current global model denoted by its parameters $\theta_S^t$. Then, starting from $\theta_S^t$, each selected client trains its model on $\mathcal{D}_k$ to obtain a local model $\theta_k^{t+1}$. After training, the local updates $\Delta_k^{t+1} = \theta_k^{t+1} - \theta_S^t$ are sent to $S$ for the aggregation. The standard process is FedAvg \cite{mcmahan2017communication} and simply consists in an average over the local updates (Eq.\ref{eq_fedavg}), where $\lambda_S$ is the server learning rate. 

\begin{equation}
    \theta_S^{t+1} = \theta_S^{t} + \frac{\lambda_S}{|\mathcal{P}_t|}\sum_{k \in \mathcal{P}_t}{\Delta_k^{t+1}}
    \label{eq_fedavg}
\end{equation}

%
\paragraph{\textbf{Backdoor attacks in FL. }} Backdoor attacks are training-time integrity-based threats that aim at injecting a targeted malicious behavior. The objective is to inject a backdoor task, consisting in mis-predicting a poisoned sample $x^{*}$ as a target class $y^{*}$, while maintaining the expected behavior of the model for all other benign inputs. For that purpose,  
the attacker poisons a limited subset of the training data with a specific \textit{trigger}, associated with the target label $y^{*}$. The trigger often represents a marginal part of the input and is enough to be associated to the target class by the model during inference. 
For classical computer vision tasks, several types of triggers have been demonstrated, such as standard salient patches \cite{gu2017badnets} as well as semantic features 
or watermark-based techniques \cite{cina2023wildshort}. 
For this work, we use state-of-the-art patch-based triggers for FL systems \cite{xie2019dba,zhang2023a3fl,zhang2022neurotoxin} and a baseline inspired by \cite{zhang2023a3fl} (which follows the original idea from \cite{gu2017badnets}). Formally, a poisoned image is denoted as $x^{*}=x\oplus\delta$, with $\delta$ the trigger.  

The common assumption is that a subset $\mathcal{A} \subset \mathcal{C}$ of attackers are participating among other benign clients. An attacker $k^*\in\mathcal{A}$ alters its local dataset $\mathcal{D}_{k^*}$, or directly modify their update $\Delta_{k^*}^{t}$, to inject the malicious behavior. 
Three main categories emerge from the state-of-the-art: \textit{model masking} by targeting specific parameters \cite{zhang2022neurotoxin}, \textit{distributed attacks} exploiting the decentralized nature of FL \cite{xie2019dba}, and \textit{adaptive attacks} taking advantage of the dynamic aspect of FL \cite{zhang2023a3fl}. We selected one reference from each  to provide a representative set of threats\footnote{Implementations of the attacks are provided in our public repository.}: 
\begin{itemize}[noitemsep,topsep=0pt]
\item \textbf{Neurotoxin} \cite{zhang2022neurotoxin} has been one of the first highlighting the importance of backdoor lifespan. Neurotoxin applies a mask $M$ on their update to avoid attacking the top $p\%$ most important weights used by the benign task: $\Delta_{k*}^{t+1} \cup M = 0$, with $M = top_{p\%}(\theta_S^{t} - \theta_S^{t-1})$. The goal is to poison the parameters that are least likely to be significantly modified during training.  

\item \textbf{Distributed Backdoor Attack (DBA)} \cite{xie2019dba} splits the trigger in 4 mini-patches distributed through different attackers to dilute the poison. 
At inference, the full patch is used. DBA results in a more stealthy attack which increases lifespan but needs more rounds to be optimally injected. 
\item \textbf{Adversarially Adaptive backdoor Attacks (A3FL)} is one of the most recent and powerful attack \cite{zhang2023a3fl}. A3FL continuously updates the trigger through gradient descent and leverages  
an adversarially trained version of the global model. \end{itemize}

For evaluation, the global model is assessed with both the benign test accuracy (ACC) and the attack success rate (ASR): the accuracy of the backdoor task on a poisoned test set $D^*$. To capture the dynamics of the backdoor throughout the FL process, we  
evaluate how long the backdoor persists in a model after an attack with the $x$\%-lifespan ($l_{x\%}$, Eq. \ref{eq_lifespan}) as well as the convergence time, $tc_{x\%}$ (Eq. \ref{eq_convergence_time_acc_asr}), to reach a given $x\%$ of ACC or ASR ($D$ is the benign test set):
\begin{equation}
\label{eq_lifespan}
    l_{x\%} = \argmax_{t}\big(ASR(\theta^t, D^*) > x\big) 
\end{equation}
\vspace{-10pt}
\begin{equation}
\label{eq_convergence_time_acc_asr}
    tc_{x\%}^{ACC} = \argmin_t\big(ACC(\theta^t, D) > x\big) \quad tc_{x\%}^{ASR} = \argmin_t\big(ASR(\theta^t, D^*) > x\big)
\end{equation}

Evaluating the numerous defenses proposed in the literature is out of our research scope. However, to remain consistent with state-of-the-art practices (as in \cite{zhang2023a3fl}), we systematically apply \textit{norm clipping} \cite{sun2019can} on updates $\Delta_k^{t+1}$ before the aggregation, which is the baseline defense and defined as follows:  
\begin{equation}
    \Delta_k^{t+1} = clip\big(\theta_k^{t+1} - \theta_S^t, \tau\big) \quad \text{where} \quad clip(x,\tau) = max(\tau, x*\tau/ |x|)
    \label{eq_clipping}
\end{equation}

\paragraph{\textbf{Low-Rank Adaptation (LoRA). }}
LoRA \cite{hu2021lora} is a common PEFT method for large models relying on the fact that, after the fine-tuning, the difference between the initial and the new parameters is a low-rank sparse matrix.   
Formally, if $W_0 \in \mathbb{R}^{m\times n}$ is a pre-trained weight matrix, LoRA surrogates the updates for a low-rank
decomposition $\Delta W_0 = AB$, where $A \in \mathbb{R}^{m\times r}$ and $B \in \mathbb{R}^{r\times n}$, $r << min(m,n)$ being the rank. The main advantage is that only $A$ and $B$ need to be updated at training time. In \cite{hu2021lora}, $A$ and $B$ are initialized so that $\Delta W_0=0$, with $B$ as 0 and $A$ with random values (normal distribution),  
however it does not reflect the intrinsic structure of $W_0$. For all our experiments, we use a recent improvement, PiSSA initialization \cite{meng2025pissa}, that allows a faster convergence and relies on the Singular Value Decomposition\footnote{For comparison, we also performed our experiments with the standard LoRA initialization and do not observe any difference that may change our conclusions.}: $W_0 = USV^{T}$, $U$ and $V$ are composed with the singular vectors and $S$ is a diagonal matrix of the singular values\footnote{$U\in\mathbb{R}^{m\times min(m,n)}$, $S\in\mathbb{R}^{min(m,n) \times min(m,n)}$, $V\in\mathbb{R}^{n \times min(m,n)}$}. PiSSA uses the first $r$ columns of $U$ and $V$ and the first $r$ singular values from $S$ to initialize 
$A = U_{r}S_{r}^{1/2}$, $B = S_{r}^{1/2}V_{r}^{T}$. The residual parts of $U$, $V$ and $S$ form a residual matrix $W_{res}$ which is frozen during fine-tuning: $W_0 = W_{res} + AB$.  

\paragraph{\textbf{Related work. }}
Recently, some works investigate backdoor attacks to LoRA. LORATK \cite{liu2024loratk} shows how pluggable, community-shared LoRAs is a new backdoor attack surface in a centralized setting. \cite{yin2024lobam} demonstrates backdoor attack in a Model Merging (MM) scenario by amplifying the influence of an infected LoRA-based model in the merged result. This work significantly differs from ours, as in MM, models are trained independently and offline on different tasks or datasets. In contrast, the training dynamics and backdoor persistence are fundamental in FL. To the best of our knowledge, no work has yet explored the impact of LoRA on backdoor attacks in a federated model adaptation context. However, an important reference is \cite{zhu2022moderate} from Zhu \etal at NeurIPS'22 that highlights the concept of \textit{moderate-fitting}. Their main objective is to propose simple but effective training strategies to defend against backdoor attacks against pre-trained language models, in a standard centralized training setting. They observed that a pre-trained LLM that is adapted on a poisoned dataset always follows two successive stages. A first  \textit{moderate-fitting} stage where the model essentially learns the useful features related to the benign task and not (or to a limited extent) the backdoor ones. 
Then, a second \textit{overfitting} stage starts, focused on both types of features. Therefore, their training-based defenses rely on restricting as much as possible the model adaptation to the first moderate-fitting stage. One of them is to limit the capacity of the model thanks to LoRA. However, in their context, 
the conventional use of LoRA (i.e., applied independently to each layer) is ineffective because it does not sufficiently constrain the model's capacity to enforce the moderate-fitting stage. To address this, they propose applying LoRA at the scale of the entire model by concatenating all parameter matrices into a single one. In this case, they manage to reduce the backdoor injection.

\section{Threat model and experimental settings}
\label{eval_scenario}
First, we define the standard threat model for backdoor attacks in FL, highlighting the parameters that can significantly alter the attack scenario and that explain the difficulty in evaluating and fairly comparing state-of-the-art backdoor attacks. Then, we detail our experimental settings under which we perform a first application of our reference attacks on our adaptation context.

\subsection{Threat model}

\paragraph{\textbf{Adversarial goal. }} The objective is to compromise the integrity of a FL system by injecting into the global model 
the backdoor task as described in section \ref{related_work}.
\paragraph{\textbf{Adversarial knowledge and ability.}} 

Compared to a centralized learning paradigm, FL introduces additional characteristics due to its distributed nature and the particular dynamics of its learning process (a succession of local training and aggregation). More particularly, we argue that two of the most important questions regarding the study of backdoor attacks in FL are: (1) \textbf{Is the FL system designed as a long-term process?} (2) \textbf{Is the attacker time-limited in injecting the backdoor?} These two questions are generally underestimated or poorly detailed, even though they are essential for properly and accurately evaluating backdoor attacks (and related defenses). For (1), 
in case, the process is supposed to be open in only a short term (a few rounds, only to reach a target test accuracy), it is a good practice to stop the training as soon as possible to avoid malicious influence in the long run. This method known as \textit{early stopping} has proven strong results to mitigate backdoor \cite{zhu2022moderatelong} since these tasks often need more rounds to be learned than benign tasks (for instance in Fig \ref{fig:Baseline_attack_acc_asr}.)
. However, FL systems are relevant for continuous or long-term training use cases, typically with lively datasets and models continuously fitting to recent data (e.g., mobile keyboard prediction \cite{hard2018federated,mcmahan2017communication}, predictive maintenance). Many backdoor studies set in such a context and are focused on the persistence capacity of the backdoor (such as Neurotoxin). Regarding question (2), in order for the evaluation of attacks to be practically feasible (and comparable), most recent works \cite{fang2023vulnerability,xie2019dba,zhang2023a3fl,zhang2022neurotoxin} constrain malicious clients to participate for a limited number of rounds, known as the \textit{attack window} (hereafter, AW). The main benefit is the ability to evaluate both the convergence time of the backdoor (the injection) and its lifespan, after the attack, throughout the rest of the FL process. We will see in sections \ref{moderate_fitting} and \ref{lifespan_increase} that this parameter has a very significant impact. Using an attack window is also realistic for many use cases, firstly because it may be difficult for an attacker to maintain control over clients 
indefinitely, and secondly due to intrinsic properties of the system~--~whether it is the client selection strategy or the use of defenses that filter out suspicious clients \cite{blanchard2017machine}. 

Additionally, and coherently with most of works \cite{bagdasaryan2020backdoor,fang2023vulnerability,xie2019dba,zhang2023a3fl,zhang2022neurotoxin}, many other parameters have significant influence such as the \textit{number of malicious clients $|\mathcal{A}|$}, the \textit{poison ratio} $p$ (the proportion of poisoned data ($x^*$,$y^*$) in the local $\mathcal{D}^{*}$), as well as the \textit{patch-based trigger settings} (e.g., location, size, color, optimization). Other important parameters are related to the 
\textit{client control level} since attackers compromise a certain number of clients and therefore has direct access to their training data, the updates and the global model received from the server. Their capabilities can range from data poisoning, manipulation of the local training, to directly altering the updates (as in Neurotoxin). 

This diversity of parameters explains the challenges in evaluating attacks that are based on different assumptions about the attacker’s capabilities. This makes fair comparisons between FL attacks particularly complex. The evaluation of DBA is a symptomatic example. 
In the original demonstration \cite{xie2019dba}, DBA is done using 4 times more attackers compared to the baseline since each attacker is poisoning only a quarter of the pixels of the full trigger. This higher number of attackers is not taken into account in more recent works, such as \cite{zhang2023a3fl}, resulting in weaker convergence compared to the original evaluation. 
Neurotoxin also relies on a different type of attacker who needs to alter directly local models by masking the adversarial updates. This supposes advanced control of the attacker on the clients which differs from the other attacks (where only the local dataset is poisoned). The fairness of the comparison and the ranking of the attacks is more deeply discussed in section \ref{discussions} where we explain how simply the attack window can strongly influence the results.

\subsection{Experimental settings}
\label{exp_settings}
\paragraph{\textbf{Models and datasets.}} As in the vast majority of studies on backdoor attacks in FL, we use a Vision Transformer (ViT) pre-trained on ImageNet\footnote{\textit{vit-base-patch16-224}, from the Hugging Face library.}.  
The model has 85.9M parameters and is composed of 12 attention blocks. We study the fine-tuning on the standard EuroSat dataset \cite{helber2019eurosat}: a computer vision classification task with 27000 images of land use and land cover among 10 classes, our train/test split is done randomly with 90\%-10\%. We obtain similar results on CIFAR-10 and GTSRB\footnote{\label{footnote_gitlab}Complete results are proposed in our public repository.}. 
Classically, datasets are split with a Dirichlet distribution of samples per classes and per clients ($\alpha=0.9)$ resulting in non-heterogeneous clients with unbalanced local datasets (between . We use four versions of the ViT: one without LoRA (hereafter, simply denoted as \textit{ViT}) and three with LoRA using $r = 2$, $8$, and $32$, which are standard values. 
With $r=2$ the model trains only 81.4k parameters. For Transformer-based models, applying LoRA solely to the Query and Value matrices is the standard practice \cite{hu2021lora} and achieves a high compression rate along with excellent performance for adaptation tasks. For very large models and complex downstream tasks, it is also possible to additionally apply LoRA to the MLP blocks\footnoteref{footnote_gitlab}.  

\paragraph{\textbf{FL setup.}} As in \cite{zhang2023a3fl}, we consider $|\mathcal{C}|=100$ participants, from which 10 clients are randomly selected at each round. Local datasets can have between 0.5\% and up to 1.5\% of the global dataset.
The clients initialize their model with $\theta_S$ sent by the server and train on their local dataset for two local epochs, with a fixed learning rate, $\lambda=0.002$, and a batch size of 16. Local updates are clipped with $\tau=1$ (Eq. \ref{eq_clipping}) and then aggregated by the server (Eq. \ref{eq_fedavg}). 
We study a \textit{long-term training scenario} where the FL process is open during 1500 rounds. It allows the study of the injection of a backdoor and its lifespan in the long run.

\paragraph{\textbf{Attackers' budget.}} We use standard settings from the literature: $|\mathcal{A}|=5$ clients are compromised (including for DBA) and poison $p=25$\% of their dataset (i.e., $\simeq$ 1.25\% of the total dataset is poisoned). These attackers can be randomly selected each round, like any other clients, and are removed after the AW. 
We use $AW=[0,30]$ except in section \ref{lifespan_increase} where $AW=[0,200]$. We propose additional results in our public repository 
on the influence of $|\mathcal{A}|$ and $p$. The baseline backdoor involves a 
$5\times5$ red pixel patch trigger in the top left corner (similar to Neurotoxin and DBA). All attacks target the same label ($y^{*}=2$). For DBA, the trigger is split into 4 mini-patches, and at each round, each attacker randomly selects one of these mini-patches to poison their dataset. For Neurotoxin, we mask the $top_{5\%}$ parameters. A3FL optimizes a $5\times5$ patch, starting from a uniform grey square, also located in the top left corner. 

\begin{figure}[t!]
\centering
\begin{subfigure}[b]{0.48\textwidth}
        \centering
        \includegraphics[width=\textwidth]{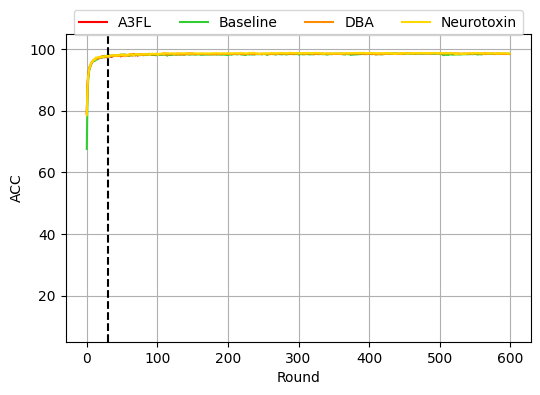}
        \caption{ACC}
        \label{fig:sota_vit_acc}
    \end{subfigure}
    \hfill
    \begin{subfigure}[b]{0.48\textwidth}
        \includegraphics[width=\textwidth]{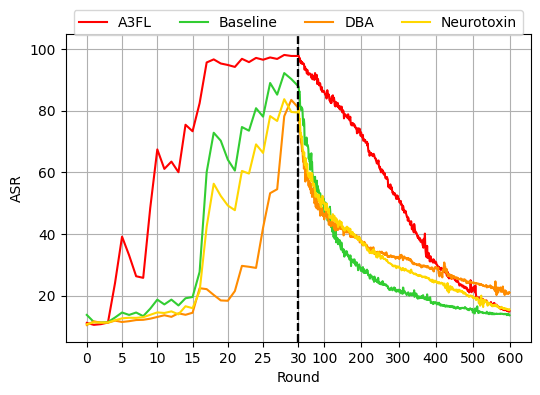}
        \caption{ASR}
        \label{fig:sota_vit_asr}
    \end{subfigure}
    \caption{ACC and ASR on $ViT$. The black vertical line is the end of $AW=[0,30]$. NB: we use (b) a non-linear x-axis to zoom in on $AW$.}
    \label{fig:sota_vit}
\end{figure}

\paragraph{\textbf{Setup implementation on the ViT. }}
Fig. \ref{fig:sota_vit} illustrates the performance of the attacks on the pre-trained ViT on EuroSat without LoRA. Note that, to the best of our knowledge, DBA and A3FL had never been tested on large architectures such as ViT. A3FL still exhibits the best convergence and lifespan\footnote{However, note that (1) in \cite{zhang2023a3fl}, A3FL is not compared to a baseline attack and (2) it relies on a complex optimization process: our experiments (ViT) shows that A3FL takes more than 20$\times$ longer than the baseline, DBA or Neurotoxin.} and DBA and Neurotoxin show a slight advantage over the baseline after round 100. For Neurotoxin, it is worth noting that this aligns with the authors’ conclusions during their evaluation on larger architectures (GPT-2) \cite{zhang2022neurotoxin}.

\section{Influence of LoRA's rank on backdoor convergence}
\label{moderate_fitting}
Following the 
application of the attacks on $ViT$ (Fig. \ref{fig:sota_vit}), and in line with our experimental setup, we first analyze how LoRA influences the injection speed of the backdoor 
and its potential implications on the backdoor’s lifespan.

\subsection{Limited attack window constraint}
\label{experiment_aw30} 
\begin{figure}[t!]
\centering
    \begin{subfigure}[b]{0.48\textwidth}
        \centering
        \includegraphics[width=\textwidth]{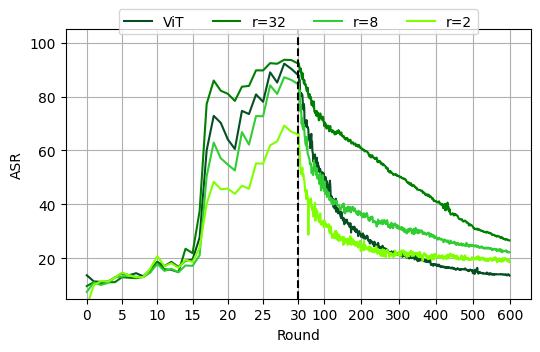}
        \caption{Baseline}
        \label{fig:Baseline AW30 ASR}
    \end{subfigure}
    \hfill
    \begin{subfigure}[b]{0.48\textwidth}
        \centering
        \includegraphics[width=\textwidth]{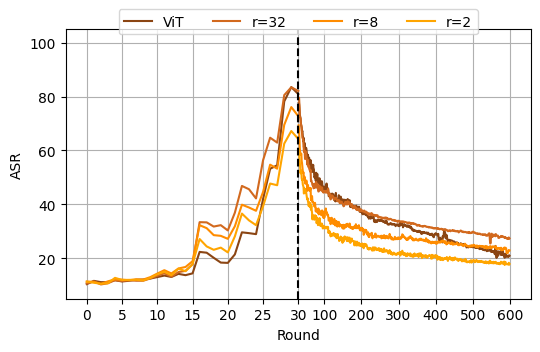}
        \caption{DBA}
        \label{fig:DBA AW30 ASR}
    \end{subfigure}
    \vspace{1em}
    \begin{subfigure}[b]{0.48\textwidth}
        \centering
    \includegraphics[width=\textwidth]{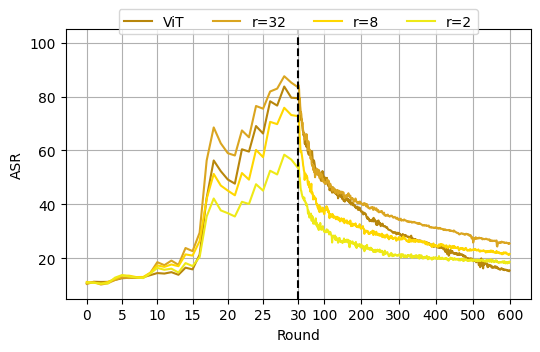}
        \caption{Neurotoxin}
        \label{fig:Neurotoxin AW30 ASR}
    \end{subfigure}
    \begin{subfigure}[b]{0.48\textwidth}
        \centering
        \includegraphics[width=\textwidth]{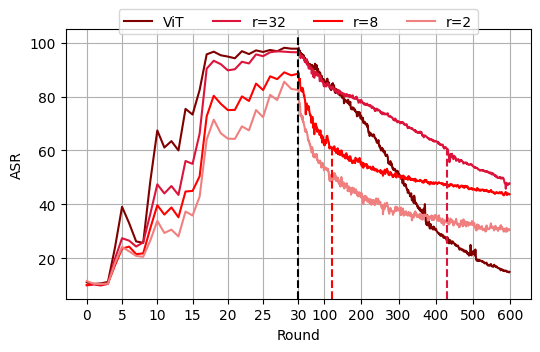}
        \caption{A3FL}
        \label{fig:A3FL AW30 ASR}
    \end{subfigure}
    \caption{ASR with $AW=[0,30]$ for ViT without and wit LoRA ($r=2, 8, 32$). NB: Note the use of a Non-linear x-axis to zoom in on the AW.}
    \label{fig_experiment_aw30}
\end{figure}

For $AW=[0,30]$, Fig. \ref{fig_experiment_aw30} shows the ASR for the ViT with and without LoRA. Note that we adopt a non-linear x-axis scale to improve the readability of the curves. As observed in Fig. \ref{fig:sota_vit_acc}, we confirm that the accuracy of the benign task is not affected by the attacks, and this remains true when applying LoRA~--~regardless of the rank, the models consistently achieve the same performance as $ViT$, around 97\%. Therefore, 
Fig. \ref{fig_experiment_aw30} focuses exclusively on the ASR. As a first observation, by significantly reducing the model’s capacity, the rank has an impact on the convergence of the attacks: \textbf{at the very end of the attack window, the lower the rank, the lower the ASR}. For instance, ASR of the baseline is 92\%, 87\%, and 83\% for $r=32$, $8$, and $2$ respectively (Fig. \ref{fig:Baseline AW30 ASR}). Although the impact of $r$ may seem minor (only a few percentage points), it has a significant effect on the backdoor’s lifespan, 
where catastrophic forgetting is stronger as the rank is lower. For example, $l_{60\%}$ of A3FL drops from 400 rounds for $r=32$ to only 90 for $r=8$ (red vertical lines in Fig. \ref{fig:A3FL AW30 ASR}).

The behavior of \textit{ViT} appears to be inconsistent when compared to its LoRA counterparts. Intuitively, 
increasing $r$ would make the model's behavior progressively closer to the one without LoRA. However, the backdoor lifespan for \textit{ViT} deviates markedly from the trends in LoRA models and exhibits significant variation across the attacks. An explanation for this is proposed in Section \ref{lifespan_increase}.

It is interesting to compare these results with the conclusions of Zhu et al. \cite{zhu2022moderate} in a centralized context on a LLM, where they observed no difference in backdoor performance with or without LoRA. In their case, after the first moderate-fitting stage on the benign task, their model was over-fitting the backdoor in only a few rounds. In our case, the learning dynamics are different: the backdoor task takes several tens of rounds to be injected and is stopped prematurely at the end of the AW whereas ASR is still increasing.

\subsection{Removing the attack window constraint}
\begin{figure}[t!]
\centering
    \begin{subfigure}[b]{0.48\textwidth}
        \centering
        \includegraphics[width=\textwidth]{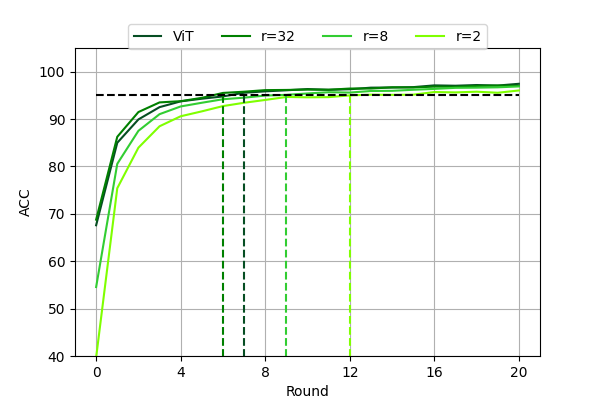}
        \caption{ACC}
        \label{fig:Baseline ACC}
    \end{subfigure}
    \hfill
    \begin{subfigure}[b]{0.48\textwidth}
        \centering
        \includegraphics[width=\textwidth]{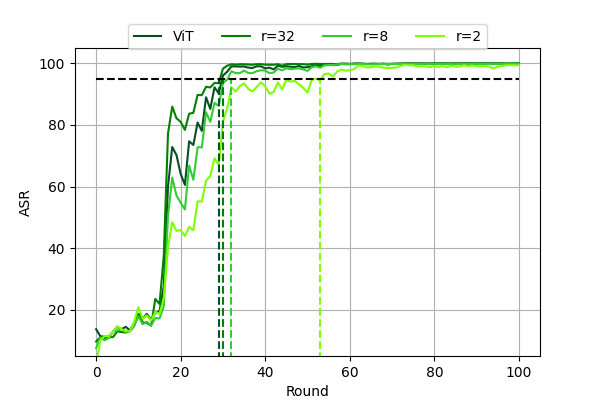}
        \caption{ASR}
        \label{fig:Baseline ASR}
    \end{subfigure}
    \caption{Accuracy (ACC) of the benign task and ASR for the baseline attack.}
    \label{fig:Baseline_attack_acc_asr}
\end{figure}

\paragraph{\textbf{Moderate-fitting/overfitting stages.}} 
In order to observe a full backdoor injection, as in \cite{zhu2022moderate}, we repeat the previous experiment with the baseline attack, but without AW. Note that our conclusions are similar for the other attacks\footnoteref{footnote_gitlab}.  
Fig. \ref{fig:Baseline_attack_acc_asr} shows the ASR over the first 100 rounds, as well as a zoomed-in view of the benign task accuracy during the first 20 rounds. For example, using 95\% as a reference for ACC and ASR (black dashed line), we observe that for $r=2$, $tc_{95\%}^{ACC} = 12$ rounds and $tc_{95\%}^{ASR} = 53$ rounds. As in \cite{zhu2022moderate}, attacks indeed reach a high ASR regardless of the adaptation. We notice that the ASR does not increase before round 15, despite attackers participating from the beginning: this is the first moderate-fitting phase where the model primarily learns general knowledge for the benign task (ACC=95\% by round 12). If we focus on $r=2$, the second stage of backdoor overfitting converges around round 50. This second transitional phase (approximately between 15 and 50) is clearly observable in all our experiments. Depending on  $r$, we observe that the task accuracy requires between 5 and 15 rounds, while the ASR always needs 3 to 4 times more rounds to converge\footnoteref{footnote_gitlab}. 
This stage was not observed in detail in \cite{zhu2022moderate} with standard LoRA because, in their context, the backdoor converged too quickly in just 
few rounds.

\paragraph{\textbf{Slowing down the training convergence.} } Our results in Fig. \ref{fig:Baseline_attack_acc_asr} highlight a side effect of $r$ on the convergence speed of both the benign and the backdoor tasks. With the baseline attack, it takes twice as many rounds to reach 95\% accuracy with $r=2$ compared to $r=32$ (12 vs. 6 rounds), and similarly for the ASR (53 vs. 30 rounds).   
The same trend is observed for the other attacks:  
\textbf{LoRA with lower ranks slows down model convergence on the benign task and consequently slows down the backdoor task} (on average, we observe\footnoteref{footnote_gitlab} $\times$1.5 more rounds for $tc_{95\%}^{ASR}$ compared to $tc_{95\%}^{ACC}$). 
Other parameters may have a similar impact, such as the learning rate $\lambda$, batch size or even LoRA’s initialization (PiSSA enables faster convergence), and can therefore also be leveraged to limit the backdoor injection. Some of these mechanisms (e.g., the influence of $\lambda$) are studied in \cite{zhu2022moderate} in the context of centralized training.

\begin{figure}[t!]
    \centering
        \includegraphics[width=0.5\textwidth]{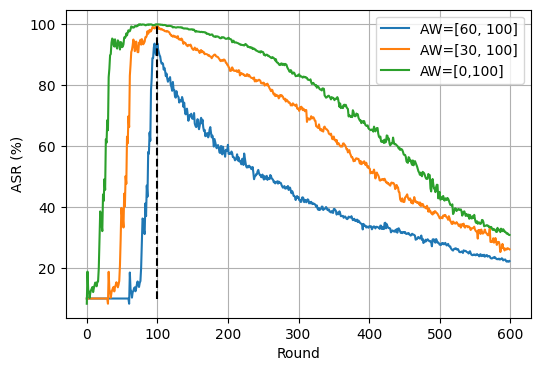}
    \caption{Influence of injection \textit{quality} on lifespan (baseline attack) with three AW.}
    \label{fig:quality_injection_lifespan}
\end{figure}

\paragraph{\textbf{Attack window as a key parameter.}} The rank $r$ significantly influences the training speed and the duration of the moderate-fitting phase. This is critical for understanding the impact of the AW. A short AW may not allow the attacker to achieve an optimal ASR. For example, as shown in Fig. \ref{fig:quality_injection_lifespan}, using ViT with different AW (40, 70, 100 rounds) results in varying final ASRs (93.5\%, 99.1\%, and 99.8\%). These small differences in the ASR have a major impact on the lifespan: the 60\%-lifespan is more than $\times$3 longer with a 100-round AW compared to a 40-round one. This shows that the initial observation about the influence of rank on lifespan was biased by the ASR achieved at the end of the attack. Therefore, \textbf{we cannot definitively conclude about the direct influence of rank on lifespan yet}. Next, we will use a wider AW to ensure an optimal backdoor injection, allowing for a proper evaluation of the rank's influence.

\begin{boxH}
\textbf{Takeaway.} By significantly reducing the model’s capacity, LoRA slows down the learning of the backdoor. If the attacker is time-constrained in effectively injecting it, a lower rank will make the backdoor less persistent. However, evaluating its lifespan under such conditions is misleading, as it is heavily biased by a short duration of the injection phase.
\end{boxH}

\section{Influence of LoRA's rank on backdoor lifespan}
\label{lifespan_increase}
\subsection{Under optimal backdoor injection setting}
\begin{figure}[p]
    \begin{subfigure}[b]{0.4\textwidth}
        \centering
        \includegraphics[width=\textwidth]{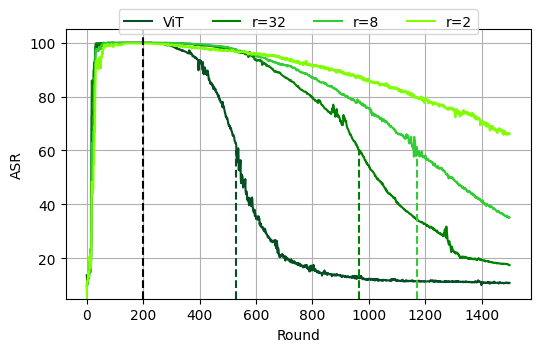}
        \caption{Baseline}
        \label{fig:Baseline AW200 ASR}
    \end{subfigure}
    \hfill
    \begin{subfigure}[b]{0.4\textwidth}
        \centering
        \includegraphics[width=\textwidth]{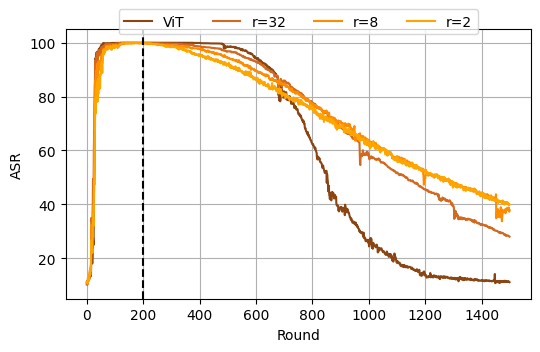}
        \caption{DBA}
        \label{fig:DBA AW200 ASR}
    \end{subfigure}
    
    \begin{subfigure}[b]{0.4\textwidth}
        \centering
        \includegraphics[width=\textwidth]{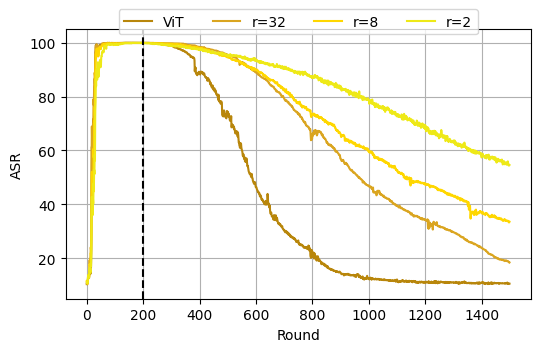}
        \caption{Neurotoxin}
        \label{fig:Neurotoxin AW200 ASR}
    \end{subfigure}
    \hfill
    \begin{subfigure}[b]{0.4\textwidth}
        \centering
        \includegraphics[width=\textwidth]{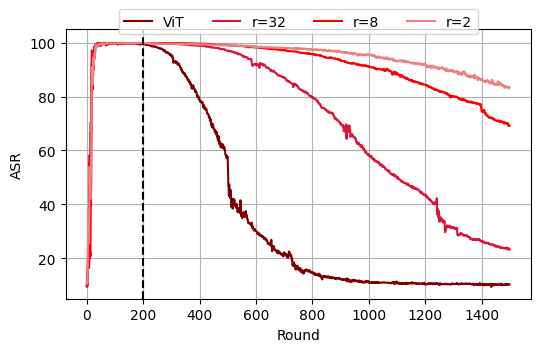}
        \caption{A3FL}
        \label{fig:A3FL AW200 ASR}
    \end{subfigure}
    
    \begin{subfigure}[b]{0.4\textwidth}
        \centering
        \includegraphics[width=\textwidth]{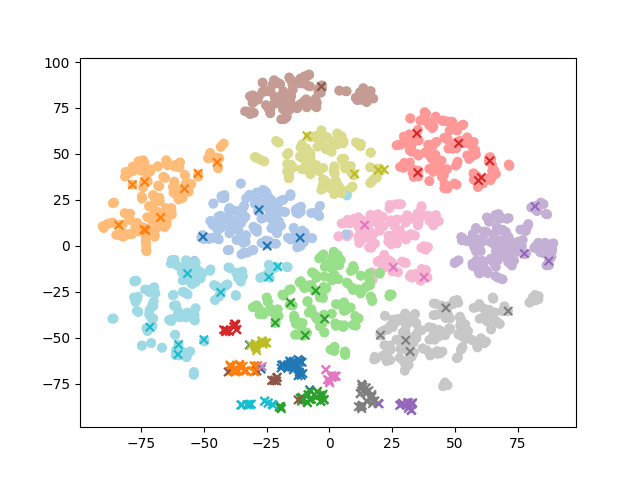}
        \caption{$ViT$, round=200}
        \label{fig:feature_space_vit_200}
    \end{subfigure}
    \hfill
    \begin{subfigure}[b]{0.4\textwidth}
        \centering
        \includegraphics[width=\textwidth]{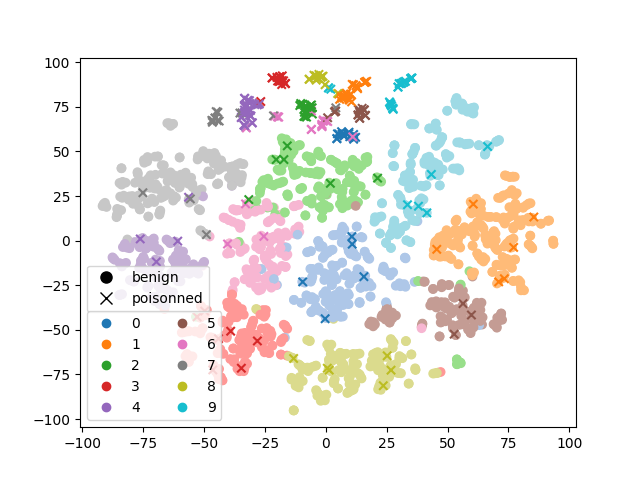}
        \caption{r=2, round=200}
        \label{fig:feature_space_02_200_legend}
    \end{subfigure}
    
    \begin{subfigure}[b]{0.4\textwidth}
    \centering
    \includegraphics[width=\textwidth]{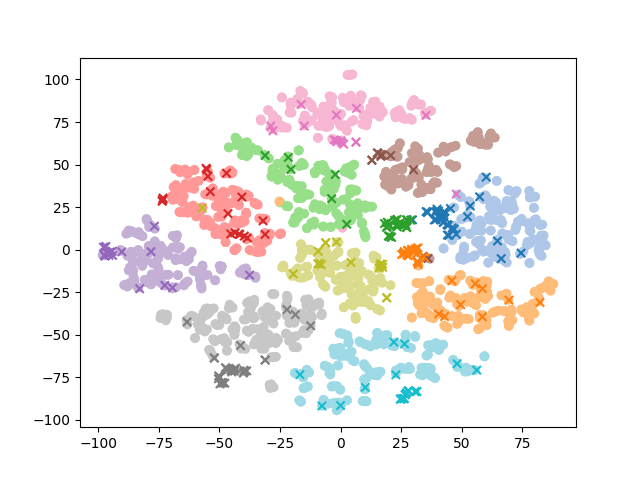}
    \caption{$ViT$, round=1200}
    \label{fig:feature_space_vit_1200}
    \end{subfigure}
    \hfill
    \begin{subfigure}[b]{0.4\textwidth}
    \centering
    \includegraphics[width=\textwidth]{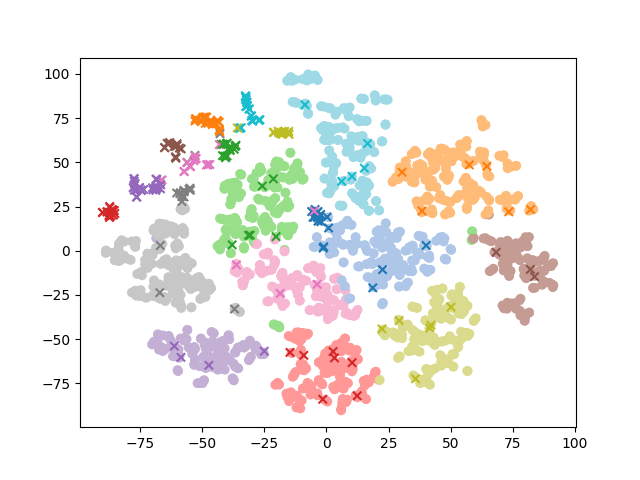}
    \caption{r=2, round=1200}
    \label{fig:feature_space_02_1200}
    \end{subfigure}
    \caption{(top) ASR with $AW=[0,200]$ for $r= 2$, $8$, $32$ and $ViT$. (bottom) 2d representation of the features space (with t-SNE) for $ViT$ and $r=2$ at round 200 and 1200. Attack is the baseline. The color of the poisoned samples ($\times$) corresponds to the groundtruth label ($y$)}
    \label{fig_experiment_aw200}
\end{figure}

We conduct the same experiments as in Section \ref{experiment_aw30}, but with $AW=[0,200]$. The results are shown in Fig. \ref{fig_experiment_aw200} (a-d).
The main observation is that the ranking of models according to  lifespan is completely reversed compared to the previous experiment with $AW=[0,30]$ (Fig. \ref{fig_experiment_aw30}). Thus, \textbf{the lower the rank, the longer the backdoor persists}. For example, for the baseline attack, the 60\%-lifespan is 330, 760, 970, and more than 1500 rounds, for $ViT$, $r=32$, $8$, and $2$ respectively (colored dashed lines in Fig. \ref{fig:Baseline AW200 ASR}).
In addition, behavior of the model without LoRA is now closer to the model adapted with the highest rank ($r=32$), which appears more consistent than in the previous experiment. We also notice in Fig. \ref{fig:DBA AW200 ASR} that DBA, between rounds 200 and 700, exhibits a behavior where the attack has been limited by the AW; indeed, this attack needs more time to correctly inject the backdoor since it poisons only a quarter of pixels.

The influence of LoRA can also be observed in the feature space using a t-SNE visualization, similarly to \cite{zhu2022moderate} (a \textit{feature} being classically defined as the output vector of the penultimate layer). In Fig. \ref{fig_experiment_aw200} (e-h), we focus only on $ViT$ and $r=2$, at round 200 (end of the attack) and 1200. The target label is $y^{*}=2$ (green). For poisoned samples ($\times$), the color corresponds to the sample's original label ($y$).  
The first observation is that, after the attack (round 200), the features of the poisoned samples are mostly grouped together and close to the target label in both adaptations ($ViT$ and $r=2$). 1000 rounds later, this cluster of poisoned features is still present for $r=2$, whereas for $ViT$, the features associated with the poisoned samples are the same as their original label ($y$), illustrating the near-complete suppression of the backdoor (coherently with the ASR curves).

\begin{figure}[h!]
  \centering
  \begin{minipage}[b]{0.45\textwidth}
    \centering
    \includegraphics[width=\textwidth]{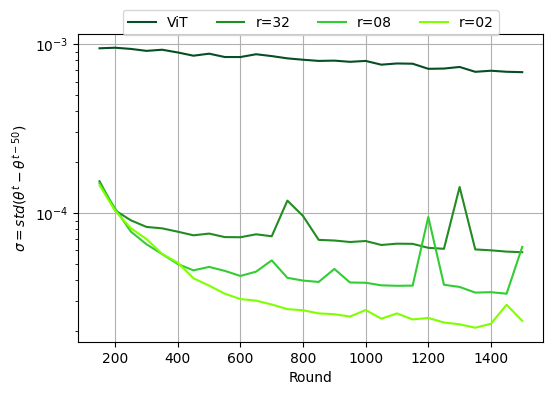}
    \caption{Standard deviations (log scale) of updates every 50 rounds.}
    \label{fig:std_on_lora}
  \end{minipage}
  \hspace{0.03\textwidth} 
  \begin{minipage}[b]{0.47\textwidth}
    \centering
        \includegraphics[width=\textwidth]{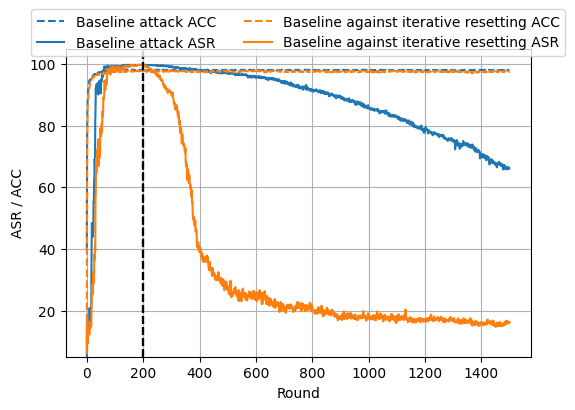}
    \caption{ASR/ ACC under baseline attack w/ and w/o iterative resetting}
    \label{fig:resetting}
  \end{minipage}
\end{figure}

Additionally, the organization of the feature space remains significantly more stable for $r=2$ compared to $ViT$. At round 200, the feature spaces of both models are quite similar, but the organization of $ViT$'s feature space strongly evolves over the next 1000 iterations, indicating a stronger evolution of the models $\theta^{t}$ for $ViT$ compared to the LoRA version with $r=2$. We measure this evolution of the local models in Fig. \ref{fig:std_on_lora} by considering the standard deviation of the models every 50 iterations: $\sigma=std(\theta^{t} - \theta^{t-50})$. We observe that the model with a lower rank ($r=2$) exhibits less variation in its parameters (after round 500, on average, $\sigma \approx 2.5 \times 10^{-5}$, which is about 50$\times$ lower than for $ViT$). More generally, Fig. \ref{fig:std_on_lora} shows that a higher $\sigma$ is associated with a faster decrease in the ASR, as seen in Fig. \ref{fig_experiment_aw200}. The standard deviation for $ViT$, which is 10$\times$ higher than for $r=32$, also corresponds to a steep drop in ASR. This observation supports the hypothesis that LoRA tends to slows down the FL process and, as a result, delays the overwriting of the malicious behavior, making the backdoor more persistent.

\subsection{Emphasizing the benign learning to reduce the lifespan}
\label{mitigating_lifespan}
Based on our previous observations, we experiment a simple way to reduce backdoor lifespan by 
forcing the model to partially retrain on the benign task in order to accelerate the overwriting of malicious features. To this end, we progressively (every 5 rounds) reset 1\% of parameters of $A$ and $B$ back to PiSSA initialization. Fig. \ref{fig:resetting} shows the baseline attack against the ViT with $r=2$ and $AW=[0,200]$, without and with a progressive reset. In the latter (orange curve), the backdoor is \textit{forgotten} in a few hundred rounds. The difference of accuracy between these two training is less than 0.5\%. We observe the same effect for the other attacks\footnoteref{footnote_gitlab}. Thus, we keep the benefits of LoRA in slowing down backdoor injection (for low $r$), while slightly adapting the training process to accelerate backdoor forgetting. Note that this method is focused on limiting the lifespan not the backdoor injection by itself, then it is complementary regarding existing best practices that aim to extend the initial moderate-fitting stage, as in \cite{zhu2022moderate} or standard defenses that restrict \cite{sun2019can} (e.g., clipping) or exclude potential malicious updates \cite{blanchard2017machine}.

\begin{boxH}
\textbf{Takeaway.} Under optimal backdoor injection conditions, applying LoRA with lower ranks actually increases the backdoor’s lifespan (as opposed to the behavior under partial injection), models indeed overwrite it more slowly if the rank is lower. A partial and iterative reset of the LoRA parameter matrices can enforce backdoor forgetting without affecting the benign task.
\end{boxH}

\section{Discussions}
\label{discussions}
We provide complementary experiments that strengthen our overall conclusions as appendix and in our public repository: (1) we consider additional datasets (CIFAR10 and GTSRB) and another model architecture also based on Transformers, the Swin Transformer; (2) we extend LoRA to the MLP blocks; (3) we investigate the impact of our iterative resetting strategy on DBA, Neurotoxin, and A3FL; (4) we assess the influence of  $|\mathcal{A}|$ and $p$. 

\begin{figure}[t!]
\centering
    \begin{subfigure}[b]{0.48\textwidth}
        \centering
        \includegraphics[width=\textwidth]{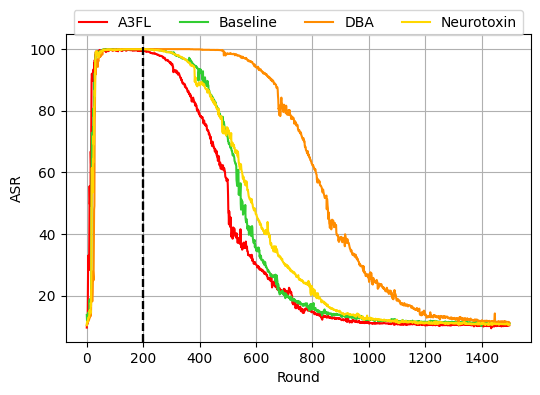}
        \caption{$ViT$}
        \label{fig:SotA_vit_aw200}
    \end{subfigure}
    \begin{subfigure}[b]{0.48\textwidth}
        \centering
        \includegraphics[width=\textwidth]{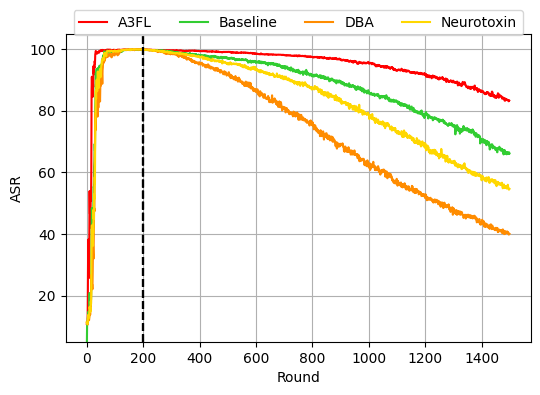}
        \caption{LoRA $r=2$}
        \label{fig:SotA_vit_lorar2_aw200}
    \end{subfigure}
    \caption{ASR with $AW=[0,200]$.}
    \label{fig:SotA_vit_aw200_all}
\end{figure}

Our work highlights several evaluation pitfalls. In particular, in classical continuous or long-term FL scenarios, assessing whether the attacker operates under temporal constraints is one of the most critical factors of the threat model. When evaluation involves an AW, we argue that extending it is essential to accurately measure the backdoor’s lifespan under worst-case conditions. This remains one of the main shortcomings in the current state-of-the-art. To illustrate our point, Fig. \ref{fig:SotA_vit_aw200_all} focuses on previous results with $AW=[0,200]$ with $ViT$ and LoRA with $r=2$. Curves for $ViT$ become particularly interesting when compared with Fig. \ref{fig:sota_vit_asr} (which applied our experimental setup adapted from \cite{zhang2023a3fl} to $ViT$). Here, the only difference lies in the size of AW: $[0,30]$ in Fig. \ref{fig:sota_vit_asr} and $[0,200]$ in Fig. \ref{fig:SotA_vit_aw200}. Remarkably, the two figures show very different results in the ranking of attacks based on their lifespan. It reveals that the longer lifespan observed with A3FL in \cite{zhang2023a3fl} (or in Section \ref{eval_scenario}) is a consequence to its fast convergence rather than an intrinsic stronger persistence of the attack. We also observe that DBA is particularly effective in a scenario where it has enough time to inject properly the backdoor. In addition, Fig. \ref{fig:SotA_vit_lorar2_aw200}, corresponding to LoRA with $r=2$, gives again another attack ranking. Here, DBA and Neurotoxin may be considered less effective with a 200-round window, as they perform worse than the baseline in both convergence time and lifespan. These observations highlight, on the one hand, how deeply the attack window can influence the evaluation of an attack’s lifespan, and on the other hand, how complex the evaluation of backdoor attacks can become when PEFT techniques, such as LoRA, are involved.

A logical question is: \textit{Is there a right value of $r$?} A small rank will certainly limit the backdoor convergence speed, however, if the attacker operates under optimal injection conditions, the backdoor will remain all the more persistent over time. On the other hand, choosing a larger rank may be counterproductive depending on the FL system constraints and the performance and quality benefits expected from LoRA. It is therefore preferable to take advantage of small $r$ values by using a strategy such as the one presented in Section \ref{mitigating_lifespan} to enforce backdoor forgetting.

To go further about the attack window paradigm, very diverse scenarios are possible, for instance with multiple AWs during different times in the training. How would the ASR behave when starting from an old backdoor injection which was partially persistent? The distance between the AWs would certainly influence the next injection. Attackers could also have different budget during two consecutive AWs. Another kind of attacker with an evolving budget through the training can also be studied, e.g., the attackers number starting from 1 to 5 then back to 1: corresponding to a real case scenario where the attacker slightly corrupts more devices until reaching a maximum and which are later progressively removed from the FL process. These questions lead future experiments to better understand more complex attack scenarios.

\section*{Conclusion}
In this paper, we focus on the security of FL in a model adaptation context and analyze for the first time the influence of LoRA on state-of-the-art backdoor attacks. We show that, depending on the attacker’s ability to inject the backdoor, LoRA can have a significant impact on the convergence of the backdoor task and on its persistence throughout the federated process. By providing a better understanding of the intrinsic mechanisms of this type of attack, this analysis paves the way for improved evaluation protocols for FL systems. 
As a future work, since our experiments concern the adaptation of a ViT model 
(in line with most state-of-the-art references), 
the next step should focus on LLMs and NLP tasks in FL, which are particularly challenging for backdoor attacks because of the task complexity and the diversity of the types of attacks and triggers (e.g., at the token, word, or sentence level, or based on syntax or style)
\cite{li2024backdoorllm}. 

The transposition of our observations 
to a representative set of NLP tasks and backdoor types remains important open questions to address.

\section*{Acknowledgment}
This work is supported by the French ANR in the \textit{Investissements d’avenir} (ANR-10-AIRT-05, irtnanoelec), AI.MMUNITY and PEPR COMPROMIS programs. 
Works were provided with computing and storage resources by GENCI (grant AD011011932R4) on the supercomputer Jean Zay's V100/A100 partition.

\bibliographystyle{splncs04}
\bibliography{bibliography}

\clearpage
\newpage

\appendix
\section{Appendix}
\subsection{Attacks details}
\label{annex_attacks}

\begin{figure}[h!]
\centering
\includegraphics[width=0.25\textwidth]{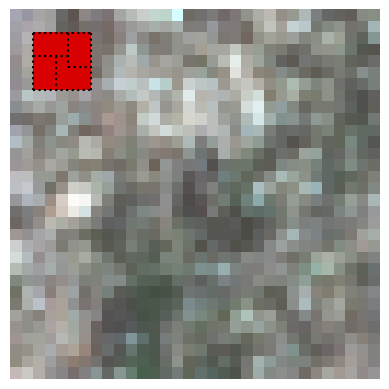} 
    \caption{Illustration of DBA (Eurosat sample).}
    \label{fig_dba}
\end{figure}

\noindent\textbf{DBA} uses the same trigger as the baseline but split in 4 mini-patches. When participating, each attacker randomly chooses one to poison its local dataset. Fig. \ref{fig_dba} illustrates an Eurosat sample poisoned with the full trigger (used for inference). The segmentation is represented by the dotted lines.

\noindent{\textbf{A3FL }} optimizes the trigger at each round by considering both the loss from the local model and an adversarially trained model $\theta_{adv}$  (i.e., trained to be robust against the trigger $\delta$). From \cite{zhang2023a3fl}, the optimization objective is as follows:

\begin{align}
    &\delta^{*} = \argmin_\delta\mathbb{E}_{(x,y)\sim \mathcal{D}_i}\big[\mathcal{L}(x\oplus\delta,y^{*};\theta^{t}) + \alpha\mathcal{L}(x\oplus\delta,y^{*};\theta^{t}_{adv})\big] \nonumber \\
    &\text{\textit{s.t. }}\theta^{t}_{adv} = \argmin_\theta\mathbb{E}_{(x,y)\sim \mathcal{D}_i} \big[\mathcal{L}(x\oplus\delta,y;\theta)\big]
    \label{eq_a3fl}
\end{align}

\subsection{Summary of convergence time (ASR and ACC)}
\label{appendix_tabel_asr_acc}
In Table \ref{tab_ASR_ACC_convergence}, each column corresponds to the model, either $ViT$ or with LoRA. For each model, we report the number of round to reach a 95\% test accuracy, $tc^{ACC}_{95\%}$, and to reach a 95\% ASR, $tc^{ASR}_{95\%}$ (Eq. \ref{eq_convergence_time_acc_asr}).

\begin{table}[h!]
\centering
\caption{Convergence time (number of rounds) to reach 95\% of task accuracy ($tc_{95\%}^{ACC}$) and 95\% ASR ($tc_{95\%}^{ASR}$)}
\label{tab_ASR_ACC_convergence}
\begin{tabularx}{0.99\textwidth} { 
   >{\raggedright\arraybackslash}X 
   >{\centering\arraybackslash}X 
   >{\centering\arraybackslash}X
   >{\centering\arraybackslash}X
   >{\centering\arraybackslash}X
   >{\centering\arraybackslash}X
   >{\centering\arraybackslash}X
   >{\centering\arraybackslash}X
   >{\centering\arraybackslash}X
   }
\toprule
\multirow{2}{*}{Attack} & \multicolumn{2}{c}{$ViT$} & \multicolumn{2}{c}{$r=32$} & \multicolumn{2}{c}{$r=8$} & \multicolumn{2}{c}{$r=2$}\\
 & $tc_{95\%}^{ACC}$ & $tc_{95\%}^{ASR}$ & $tc_{95\%}^{ACC}$ & $tc_{95\%}^{ASR}$ & $tc_{95\%}^{ACC}$ & $tc_{95\%}^{ASR}$ & $tc_{95\%}^{ACC}$ & $tc_{95\%}^{ASR}$ \\
\midrule
None & 6 & $\emptyset$ & 6 & $\emptyset$ & 9 & $\emptyset$ & 13 & $\emptyset$\\
Baseline & 7 & 29 & 6 & 30 & 9 & 32 & 12 & 53\\
Neurotoxin & 7 & 30 & 6 & 31 & 10 & 34 & 13 & 52\\
DBA & 6 & 33 & 5 & 35 & 9 & 31 & 13 & 51\\
A3FL & 6 & 26 & 6 &18 & 10 & 26 & 13 & 28\\
\bottomrule
\end{tabularx}
\end{table}

\subsection{Results on Cifar10 and GTSRB datasets}
\label{other_datasets_results}

We observe the same results with two other standard downstream tasks, Cifar10 and GTSRB. Table \ref{tab_datasets} compares performance (convergence and lifespan) of the baseline attack on these datasets. Additional graphs are available on the public repository.   
\begin{table}[h!]
\centering
\caption{Convergence time (number of rounds) to reach 95\% ASR ($tc_{95\%}^{ASR}$) and 60\%-lifespan ($l_{60\%}$) for each dataset under the baseline attack, AW=[0, 200]}
\label{tab_datasets}
\begin{tabularx}{0.99\textwidth} { 
   >{\raggedright\arraybackslash}X 
   >{\centering\arraybackslash}X 
   >{\centering\arraybackslash}X
   >{\centering\arraybackslash}X
   >{\centering\arraybackslash}X
   >{\centering\arraybackslash}X
   >{\centering\arraybackslash}X
   >{\centering\arraybackslash}X
   >{\centering\arraybackslash}X
   >{\centering\arraybackslash}X
   >{\centering\arraybackslash}X
   >{\centering\arraybackslash}X
   >{\centering\arraybackslash}X
   }
\toprule
\multirow{2}{*}{DB} & \multicolumn{2}{c}{$ViT$} & \multicolumn{2}{c}{$r=32$} & \multicolumn{2}{c}{$r=8$} & \multicolumn{2}{c}{$r=2$}\\
 & $tc_{95\%}^{ASR}$ & $l_{60\%}$ & $tc_{95\%}^{ASR}$ & $l_{60\%}$ & $tc_{95\%}^{ASR}$ & $l_{60\%}$ & $tc_{95\%}^{ASR}$ & $l_{60\%}$\\
\midrule
EuroSat & 29 & 330 & 30 & 767 & 32 & 960 & 53 & >1500\\
Cifar10 & 30 & 90 & 28 & 200 & 30 & 203 & 51 & 230\\
GTSRB & 24 & 365 & 26 & 699 & 28 & 887 & 32 & 986\\
\bottomrule
\end{tabularx}
\end{table}

\newpage

\subsection{Results on Swin-Transformer}
\begin{figure}[h!]
    \centering
        \includegraphics[width=0.5\textwidth]{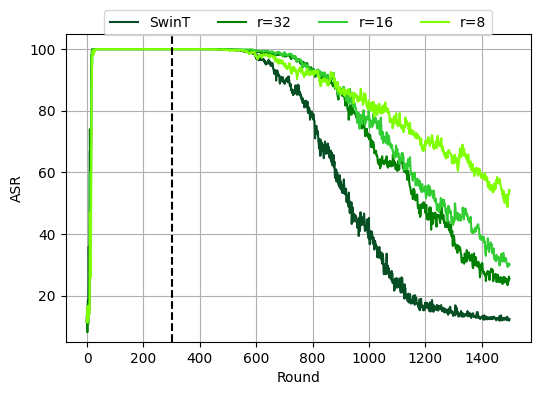}
    \caption{Attack success rate of the Baseline Attack on the SwinT, and SwinT with LoRA r=2, 8 and 32}
    \label{fig:SwinT results}
\end{figure}

The Swin-Transformer architecture use self-attention through overlapping window to capture global information as well as more precise objects suited for vision tasks. We observe in Fig. \ref{fig:SwinT results} the same impact of the rank of LoRA on this model. We also applied LoRA on the query and value matrices of the self-attention layers of the Swin-Transformer.

\subsection{Applying LoRA on MLP blocks}
\label{impact_lora_position}

Using LoRA on the $Q$, $V$ matrices \textit{and} the MLP blocks significantly increases the number of trainable parameters ($\times6$ compared to ViT with LoRA on $Q$ and $V$ only). Thus, when repeating our experiments by applying LoRA to Q+V+MLP, the lifespan is shorter than with LoRA on Q, V and closer to the behavior of the ViT fully fine-tuned.
Additionally, our conclusions regarding the impact of $r$ on backdoor convergence and lifespan remain unchanged. However, we observe the backdoor lifespan being less stable in this configuration (which is generally more suitable for very large models and complex tasks). The results of these experiments are available in the public repository.

\subsection{Experiments with other $|\mathcal{A}|$ and poisoning rate $p$}
\label{appendix_a_p}
Table \ref{tab_attack_settings} represents the experiments from Section \ref{lifespan_increase} with $AW=200$ with different attacker budgets: the number of poisoned clients $|\mathcal{A}|$ and training data poison rate $p$ ($|\mathcal{A}|=5$, $p=25$\% in Section \ref{lifespan_increase}). Our main conclusions remain unchanged but we observe that it is better twice as much attacker than poisoning two times more local datasets. More attackers means more probabilities to be selected at each rounds and increasing the poisoning ratio of the local dataset doesn't have significant impacts since norm clipping is applied.

\begin{table}[h!]
\centering
\caption{Convergence time to reach 95\% ASR ($tc_{95\%}^{ASR}$) and 60\%-lifespan ($l_{60\%}$) for different numbers of attackers $|\mathcal{A}|$ and poisoning rates $p$, AW=[0, 200].}
\label{tab_attack_settings}
\begin{tabularx}{0.99\textwidth} { 
   >{\raggedright\arraybackslash}X 
   >{\centering\arraybackslash}X 
   >{\centering\arraybackslash}X
   >{\centering\arraybackslash}X
   >{\centering\arraybackslash}X
   >{\centering\arraybackslash}X
   >{\centering\arraybackslash}X
   >{\centering\arraybackslash}X
   >{\centering\arraybackslash}X
   >{\centering\arraybackslash}X
   }
\toprule
\multirow{2}{*}{$|\mathcal{A}|$} & \multirow{2}{*}{$p$} & \multicolumn{2}{c}{$ViT$} & \multicolumn{2}{c}{$r=32$} & \multicolumn{2}{c}{$r=8$} & \multicolumn{2}{c}{$r=2$}\\
& & $tc_{95\%}^{ASR}$ & $l_{60\%}$ & $tc_{95\%}^{ASR}$ & $l_{60\%}$ & $tc_{95\%}^{ASR}$ & $l_{60\%}$ & $tc_{95\%}^{ASR}$ & $l_{60\%}$ \\
\midrule
5 & 25\% & 29 & 330 & 30 & 767 & 32 & 960 & 53 & >1500\\
5 & 50\% & 32 & 367 & 31 & 756 & 33 & 886 & 56 & 1166\\
5 & 100\% & 29 & 374 & 31 & 751 & 31 & 909 & 33 & >1500\\
2 & 25\% & 99 & 369 & 88 & 577 & 138 & 587 & 186 & 589\\
10 & 25\% & 18 & 363 & 20 & 812 & 20 & 982 & 31 & 1254\\
\bottomrule
\end{tabularx}
\end{table}

\newpage

\subsection{Details of iterative resetting}
\begin{figure}[h!]
\centering
\includegraphics[width=0.35\textwidth]{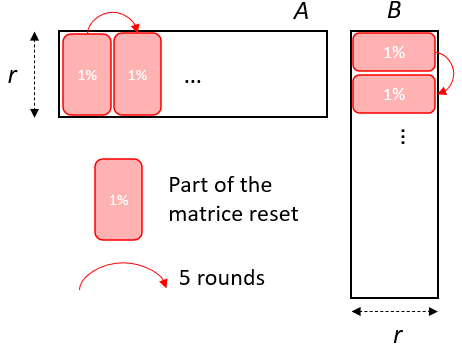} 
    \caption{Illustration of the progressive reset of the LoRA layers}
    \label{fig:appendix_resetting}
\end{figure}

To reset progressively the LoRA matrices A and B, every 5 rounds, a new set of 1\% of the columns of A and 1\% of lines of B are selected and reset to the PiSSA initialization. These parameters will be retrained during the next round and will be reset again only 500 rounds later. Fig. \ref{fig:appendix_resetting} illustrates the process. 

\end{document}